\documentclass[10pt,twocolumn,letterpaper]{article}

\usepackage[pagenumbers]{cvpr} 

\usepackage{graphicx}
\usepackage{amsmath}
\usepackage{amssymb}
\usepackage{booktabs}
\graphicspath{{./figures/}}
\usepackage{algorithm}
\usepackage{algorithmic}
\usepackage{enumitem}
\usepackage{color}
\usepackage{caption}
\usepackage{colortbl}
\usepackage{booktabs}

\makeatletter
\@namedef{ver@everyshi.sty}{}
\makeatother
\usepackage{tikz}

\definecolor{Gray}{gray}{0.9}
\definecolor{myfavblue}{rgb}{0.05, 0.2, 0.8}
\definecolor{keywords}{RGB}{255,0,90}
\definecolor{comments}{RGB}{0,0,113}
\definecolor{red}{RGB}{240,0,0}
\definecolor{gray}{RGB}{150,150,150}
\definecolor{green}{RGB}{0,170,0}
\definecolor{myblue}{HTML}{3182bd}
\definecolor{myred}{HTML}{de2d26}
\definecolor{mydarkblue}{rgb}{0,0.08,0.45}

\newcommand{\redtri}{
    \begin{tikzpicture}
    \filldraw[red] (0, 0)--(2.5pt, 5pt)--(5pt, 0pt)--cycle; 
    \end{tikzpicture}}
\newcommand{\bluetri}{
    \begin{tikzpicture}
    \filldraw[blue] (0, 0)--(2.5pt, 5pt)--(5pt, 0pt)--cycle; 
    \end{tikzpicture}}

\usepackage[pagebackref,breaklinks,colorlinks]{hyperref}

\usepackage[capitalize]{cleveref}
\crefname{section}{Sec.}{Secs.}
\Crefname{section}{Section}{Sections}
\Crefname{table}{Table}{Tables}
\crefname{table}{Tab.}{Tabs.}

\def\cvprPaperID{12832} 
\def\confName{CVPR}
\def\confYear{2023}

\begin{document}
\title{Fine-Grained Classification with Noisy Labels}

\author{Qi Wei$^1$, Lei Feng$^2$, Haoliang Sun$^{1}$\thanks{denotes corresponding author.}, Ren Wang$^1$, Chenhui Guo$^1$, Yilong Yin$^{1*}$\\
$^1$ School of Software, Shandong University, China \\
$^2$ School of Computer Science and Engineering, Nanyang Technological University, Singapore\\
{\tt\small \{1998v7, haolsun.cn\}@gmail.com, feng0093@e.ntu.edu.sg, rwang@mail.sdu.edu.cn} \\
{\tt\small guiguigh@163.com, ylyin@sdu.edu.cn}
}
\maketitle

\begin{abstract}
Learning with noisy labels (LNL) aims to ensure model generalization given a label-corrupted training set. In this work, we investigate a rarely studied scenario of LNL on fine-grained datasets (LNL-FG), which is more practical and challenging as large inter-class ambiguities among fine-grained classes cause more noisy labels. We empirically show that existing methods that work well for LNL fail to achieve satisfying performance for LNL-FG, arising the practical need of effective solutions for LNL-FG. To this end, we propose a novel framework called stochastic noise-tolerated supervised contrastive learning (SNSCL) that confronts label noise by encouraging distinguishable representation. Specifically, we design a noise-tolerated supervised contrastive learning loss that incorporates a weight-aware mechanism for noisy label correction and selectively updating momentum queue lists. By this mechanism, we mitigate the effects of noisy anchors and avoid inserting noisy labels into the momentum-updated queue. Besides, to avoid manually-defined augmentation strategies in contrastive learning, we propose an efficient stochastic module that samples feature embeddings from a generated distribution, which can also enhance the representation ability of deep models. SNSCL is general and compatible with prevailing robust LNL strategies to improve their performance for LNL-FG. Extensive experiments demonstrate the effectiveness of SNSCL.
\end{abstract}
\section{Introduction}\label{sec:introduction}
Learning from noisy labels~\cite{long2008random,bossard2014food,han2018co,xu2019l_dmi,li2020dividemix,wei2022self,sun2022learning,liu2020early,natarajan2013learning} poses great challenges for training deep models, whose performance heavily relies on large-scaled labeled datasets. Annotating training data with high confidence would be resource-intensive, especially for some domains, such as medical and remote sensing images~\cite{li2022exploring,sun2017direct}. Thus, label noise would inevitably arise and then remarkably degrade the generalization performance of deep models. 

\begin{figure}[t]
\centering
\scalebox{1.0}{
\includegraphics[width=1.0\linewidth]{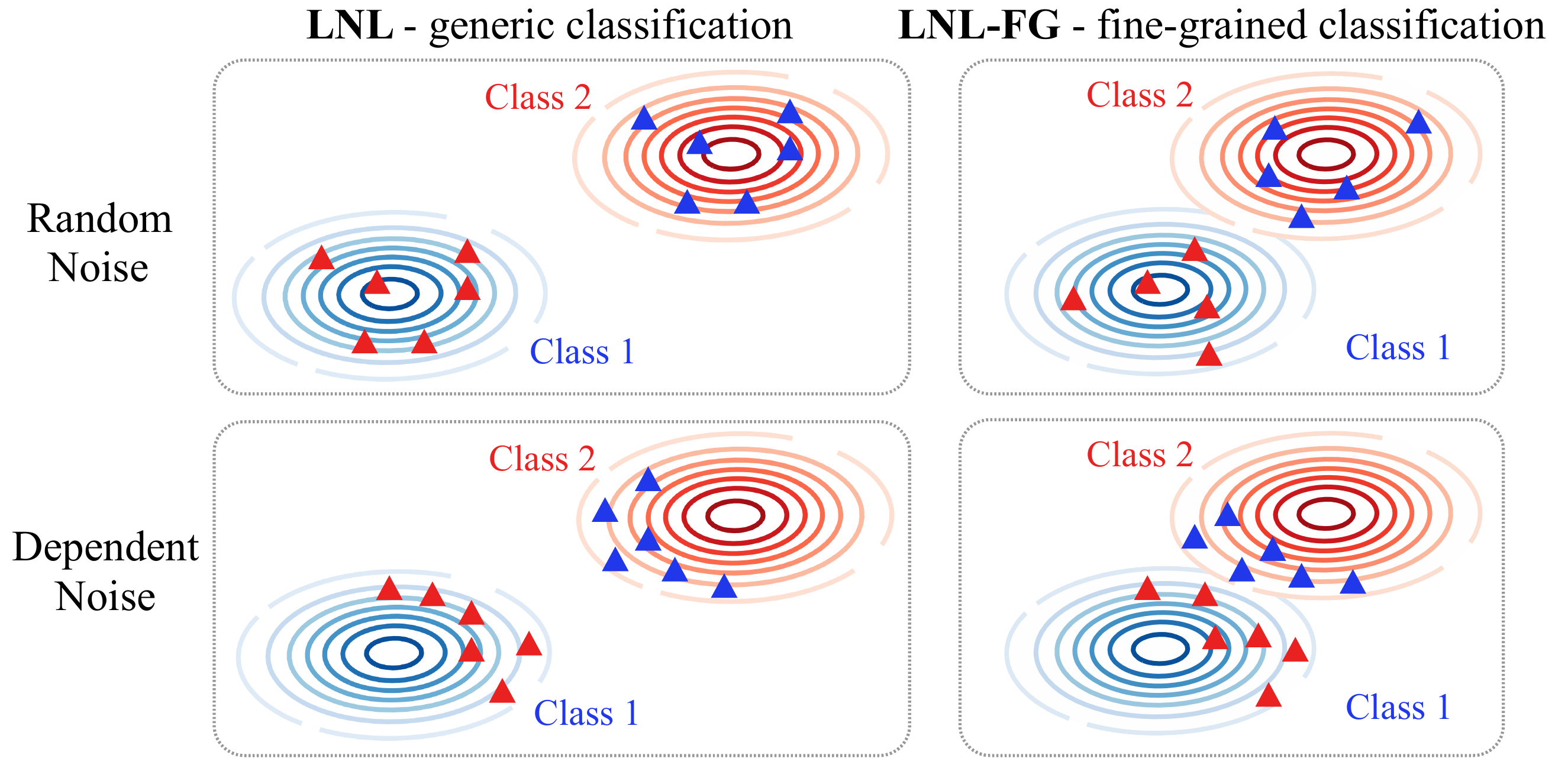}}
\vspace{-5mm}
\caption{LNL-FG is more challenging than LNL on generic classification. \protect\redtri \protect\bluetri \, denote mislabeled samples.} 
\label{fig:intro1}
\vspace{-4mm}
\end{figure} 

\begin{figure*}[t]
\centering
\includegraphics[width=1.0\linewidth]{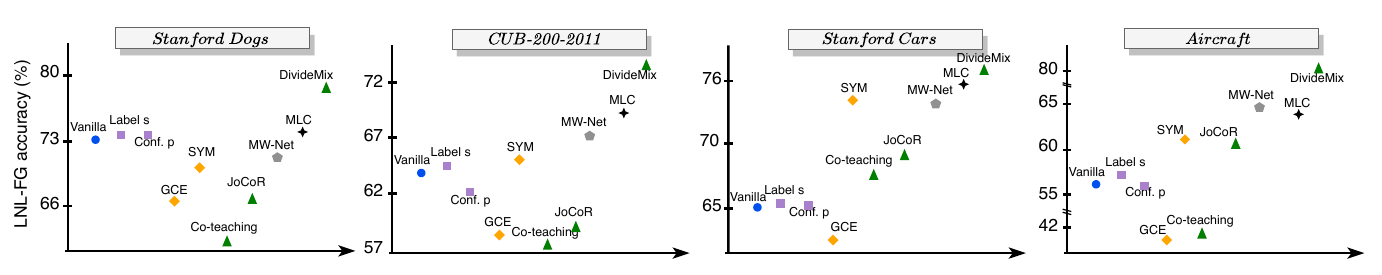}
\vspace{-9mm}
\caption{Comparison results of previous methods on four fine-grained benchmarks with 20\% random label noise. Methods with same color and shape belong to the same strategy. The \textbf{X-axis} denotes their performance on typical LNL tasks while the \textbf{Y-axis} denotes that on LNL-FG tasks. It is obvious that \textbf{not all robust methods outperform the performance of \textit{vanilla} cross-entropy on LNL-FG task.}. More analysis and results can be found in Appx. A.} 
\vspace{-3mm}
\label{fig:intro2}
\end{figure*} 

Previous methods~\cite{han2018co,bai2021me,shu2019meta,wei2020combating,li2020dividemix,long2008random,bossard2014food} in LNL always focus on generic classification (\textit{e.g.} CIFAR-10 \& 100) and artificially construct random label noise~\cite{long2008random,tewari2007consistency,tanaka2018joint,liu2020early} and dependent label noise~\cite{han2018co,wei2020combating,shu2019meta,li2020dividemix,wei2022self} to evaluate the performance of their algorithms. In this work, we extend LNL to \textit{fine-grained} classification, which is a rarely studied task. Firstly, this scenario is more realistic since annotators are easier to be misguided by indistinguishable characteristics among fine-grained images and give an uncertain target. Fig. \ref{fig:intro1} illustrates comparison between two types of noise simulated on generic and fine-grained sets. Further, we extensively investigate the performance of prevailing LNL methods on our proposed LNL-FG task. The detailed results are shown in Fig. \ref{fig:intro2}. Although these robust algorithms lead to statistically significant improvements over vanilla softmax cross-entropy on LNL, these gains do not transfer to LNL-FG task. Instead, some methods degrade the generalization performance of deep models compared to cross-entropy.
Intuitively, due to large inter-class ambiguity among those classes in LNL-FG, the margin between noisy samples and the decision boundary in the fine-grained dataset is smaller than that in the generic dataset, leading to severe overfitting of deep models to noisy labels. 
From this perspective, we consider that encouraging discrimitive feature not only confronts overfitting to label noise but also facilitates the learning of fine-grained task.

For this, contrastive learning (CL), as a powerful unsupervised learning approach for generating discrimitive feature \cite{chen2020simple,he2020momentum,grill2020bootstrap,jaiswal2020survey,park2020contrastive}, has attracted our attention. CL methods usually design objective functions as supervised learning to perform pretext similarity measurement tasks derived from an unlabeled dataset, which can learn effective visual representations in downstream tasks, especially for fine-grained classification\cite{bukchin2021fine}. The following work, supervised contrastive learning (SCL) \cite{khosla2020supervised}, leverages label information to further enhance representation learning, which can avoid a vast training batch and reduce the memory cost. However, SCL cannot be directly applied to the noisy scenario as it is lack of noise-tolerated mechanism.

To resolve the noise-sensitivity of SCL, we propose a novel framework named stochastic noise-tolerated supervised contrastive learning (SNSCL), which contains a noise-tolerated contrastive loss and a stochastic module.
For the noise-tolerated contrastive loss, we roughly categorize the noise-sensitive property of SCL into two parts of noisy anchors and noisy query keys in the momentum queue. To mitigate the negative effect introduced by noisy anchors or query keys, we design a weight mechanism for measuring the reliability score of each sample and give corresponding weight. Based on these weights, we modify the label of noisy anchors in current training batch and selectively update the momentum queue for decreasing the probability of noisy query keys. These operations are adaptive and can achieve a progressive learning process. 
Besides, to avoid manual adjustment of strong augmentation strategies for SCL, we propose a stochastic module for more complex feature transformation. In practice, this module generates the probabilistic distribution of feature embedding. By sampling operation, SNSCL achieves better generalization performance for LNL-FG. 

Our contributions can be summarized as
\begin{itemize}[leftmargin=6mm]
\setlength\itemsep{0mm}
\vspace{-2mm}

\item We consider a hardly studied LNL task, dubbed LNL-FG and conduct empirical investigation to show that some existing methods in LNL cannot achieve satisfying performance for LNL-FG.

\item  We design a novel framework dubbed stochastic noise-tolerated supervised contrastive learning (SNSCL), which alters the noisy labels for anchor samples and selectively updates the momentum queue, avoiding the effects of noisy labels on SCL.
\item  We design a stochastic module to avoid manually-defined augmentation, improving the performance of SNSCL on representation learning.
\item  Our proposed SNSCL is generally applicable to prevailing LNL methods and significantly improves their performance on LNL-FG.
\vspace{-1.5mm}
\end{itemize}

Extensive experiments on four fine-grained datasets and two real-world datasets consistently demonstrate the state-of-the-art performance of SNSCL, and further analysis verify its effectiveness.


\section{Related Work}\label{sec:related}
\vspace{-1.5mm}
\textbf{Robust methods in Learning with noisy labels}. The methods in the field of learning with noisy labels can be roughly categorized into robust loss function, sample selection, label correction, and sample reweight. The early works~\cite{zhang2018generalized,wang2019symmetric,ma2020normalized,liu2020peer} mainly focus on designing robust loss functions which provide the deep model with greater generalization performance compared with the cross-entropy loss and contain the theoretical guarantee~\cite{ma2020normalized,liu2020peer}. Currently, more works turn to explore the application of the other three strategies. In label correction, researchers refurbish the noisy labels by self-prediction of the model's output~\cite{song2019selfie,wang2021proselflc} or an extra meta-corrector~\cite{wu2020learning,zheng2021meta}. The latter enables admirable results of correction with a small set of meta-data. In sample section, the key point is how effective the preset selection criterion is. Previous literatures leverage the small-loss criterion that selects the examples with small empirical loss as the clean one~\cite{han2018co,wei2020combating}. Recently, the works~\cite{nguyen2019self,bai2021me,wei2022self} represented by SELF~\cite{nguyen2019self} pay more attention to history prediction results, providing selection with more information and thus promoting the selection results. Besides, sample reweight methods~\cite{shu2019meta,ren2018learning} give examples with different weights, which can be regarded as a special form of sample selection. For example, \cite{shu2019meta} designed a meta-net for learning the mapping from loss to sample weight. The samples with large losses are seen as the noise, and thus meta-net generates small weights.

\textbf{Contrastive learning}. As an unsupervised learning strategy, contrastive learning~\cite{chen2020simple,chen2020big,he2020momentum} leverages similarity learning and markedly improves the performance of representation learning. The core idea of these methods is maximizing (minimizing) similarities of positive (negative) pairs at the data points. 

CL has also been applied to LNL field for better representation learning and tackle negative effects of noisy labels. Sel-CL \cite{li2022selective} proposes a pair-wise framework of selecting clean samples and conducts contrastive learning on those samples. Our proposed NTSCL is different in three aspects: 1) a different selection strategy via a novel weight-aware mechanism; 2) a stochastic module avoiding manually-defined augmentations in SCL for LNL.
3) a plug-and-play module for typical LNL methods. Our method NTSCL can be easily integrated into existing methods for improving performance on LNL or LNL-FG, while Sel-CL cannot. Besides, li \textit{et al.} \cite{li2021mopro} introduces the ideas of momentum prototypes and trains the network such that embeddings are pulled closer to their corresponding prototypes, while pushed away from other prototypes. Due to the large inter-class ambiguity in fine-grained datasets, the quality  of constructed class prototypes may be challenged.

\section{Preliminaries}\label{sec:pre}
\begin{figure*}[t]
\centering
\includegraphics[width=0.62\linewidth]{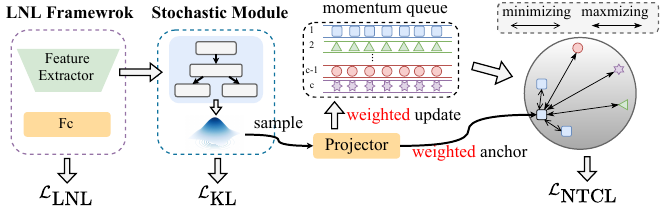}
\vspace{-3mm}
\caption{\textbf{Illustration of training framework}. Examples in the momentum queue with the same color and shape belong to the same category. The \textit{Projector} is set as a single-layer MLP structure. Overall, the total training framework includes a LNL method and our proposed SNSCL, which consists of two parts: 1) \textbf{stochastic module}, which provides more competitive feature transformation for contrastive learning; 2) \textbf{noise-tolerated contrastive loss}, which is noise-aware and contains two weighting strategies.} 
\vspace{-3mm}
\label{fig:all_fig}
\end{figure*}

\noindent\textbf{Problem definition}. Assume $\mathcal{X}$ is the feature space from which the examples are drawn, and $\mathcal{Y} = \{1,2,\cdot\cdot\cdot,C\}$ is the class label space, \textit{i.e.} we consider a $C$-classification problem. Given a training set $\mathcal{D}^N = \{({\bf x}_i, {\rm y}_i)\}_{i=1}^n $ with partial corrupted labels, where $({\bf x}_i, {\rm y}_i)$ is drawn $\rm {i.i.d.}$ according to an distribution, over $(\mathcal{X}, \mathcal{Y})$.
Supposing there is a deep classification network $F(\cdot)$ with the learnable parameters $\theta$. For sample $\bf x$, the model's output can be written as $F({\bf x}, {\bf{\theta}})$.

The goal of our algorithm is finding the optimal parameter $\theta^*$ which can achieve admirable generalization performance on the clean testing set.

\noindent\textbf{Contrastive learning meets noisy labels}. \textit{Contrastive learning} ~\cite{chen2020simple,he2020momentum,grill2020bootstrap} is a prevailing framework for representation learning, enhancing class discrimination of the feature extractor. Supposing a feature anchor $\mathrm{q}$ and a set of feature keys $\{ \mathrm{\hat q}, \mathrm{k_1},\cdot \cdot \cdot,\mathrm{k_D} \}$ are given, where $\mathrm{\hat q}$ is a positive data point for $\mathrm{q}$, and the others are negative.  In CL, a widely used loss function for measuring the similarity of each data point is InfoNCE~\cite{oord2018representation} and can be summarised as 
{\small
\begin{equation}\label{eq:L_cl}
    \mathcal{L}_{\rm INFO} = - \log \frac
    {\exp{(\mathrm{q} \cdot \mathrm{\hat q} / \tau)}}
    {\exp{(\mathrm{q} \cdot \mathrm{\hat q} / \tau)} + 
    \sum\nolimits_{d=1}^{\mathrm{D}} \exp{(\mathrm{q} \cdot \mathrm{k}_d / \tau)}},  \nonumber
\end{equation}
}
where $\tau$ is a hyper-parameter for temperature scaling. 
In most applications, CL is built as a pre-task. $\mathrm{q}$ and $\mathrm{\hat q}$ are extracted from two augmented views of the same example, and negative keys $\{\mathrm{k_1},\cdot \cdot \cdot,\mathrm{k_D}\}$ represent feature embeddings of other samples in the current training batch.
CL is naturally independent of noisy labels, but there exists a drawback in that it lacks a mechanism to utilize potential labels into model training, leaving useful discriminative information on the shelf~\cite{wang2021self}. Currently, \textit{supervised contrastive learning} \cite{khosla2020supervised} solves this issue by constructing the positive and the negative lists according to the labels. For anchor point $\mathrm{q}$, the objective function can be written as
{
\small
\begin{equation}\label{eq:L_SCL}
    \mathcal{L}_{\rm SCL} = - \log \frac
    {\sum\limits_{\mathrm{k_P} \in {\rm Pos}} \exp{(\mathrm{q} \cdot \mathrm{k_P} / \tau)}}
    {
    \sum\limits_{\mathrm{k_P} \in {\rm Pos}} \exp{(\mathrm{q} \cdot \mathrm{k_P} / \tau)} + 
    \sum\limits_{\mathrm{k_N} \in {\rm Neg}} \exp{(\mathrm{q} \cdot \mathrm{k_N} / \tau)}}, \nonumber
\end{equation}
}
where ${\rm Pos}$ and ${\rm Neg}$ represent the positive and negative list, respectively.

However, SCL is sensitive to noisy labels, which can be introduced into the anchor point, ${\rm Pos}$, and ${\rm Neg}$. Our goal is to utilize the valuable information of the labels underlying the noisy training set $\mathcal{D}^N$ and overcome the misguidance of noisy labels.

\section{Proposed method}\label{sec:method}
\noindent\textbf{Overview}. In section \ref{sec:method_1}, we first introduce a noise-tolerated supervised contrastive learning method that incorporates a weight-aware mechanism for measuring the reliability score of each example. Based on this mechanism, we dynamically alter the unreliable labels and selectively insert them into the momentum-updated queue, combating two noise-sensitive issues of SCL, respectively. Then, in section \ref{sec:method_2}, we design a stochastic module for the transformation of feature embeddings, which samples from a generated probabilistic distribution. Eventually, we exhibit the total training objective in section \ref{sec:method_3}.

\subsection{Noise-tolerated supervised contrastive learning}
\label{sec:method_1}
\noindent\textbf{Weight-aware mechanism}. We aim to measure the reliability score of each sample in the training set $\mathcal{D}^N$ and generate the corresponding weight. For this, we use the \textit{small-loss} criterion, a common strategy in LNL, and leverage a two-component GMM to generate this reliability score. Firstly, we evaluate the training set $\mathcal{D}^N$ after each training epoch. For clarity, we omit the epoch sequence and attain a list of empirical losses $\{l_i\}_{i=0}^n$ among all samples, where $l_i = L(F({\bf x}_i;\theta), {\rm y}_i)$. Note that $L(\cdot)$ is the employed loss function. GMM fits to this list and gives the reliability score of the probability that the sample is clean. For sample $x_i$, the reliability score $\gamma_i$ can be written as 
$\gamma_i = {\rm GMM} (l_i \, |\, \{l_i\}_{i=0}^n)$, 
where $\gamma_i \in [0,1]$. Then, we design a function to dynamically adjust the weight for all training samples according to the reliability score. The weight of sample ${\bf x}_i$ is
\begin{equation}\label{eq:eq3}
    \begin{aligned}
        \omega_i = 
        \begin{cases}
        1 & \text{if} \quad \gamma_i > t\\
        \gamma_i &  \text{otherwise} \\
        \end{cases},
    \end{aligned}
\end{equation}
where $t$ is a hyper-parameter in the interval of $[0,1]$ and denotes the threshold of the reliability score. The computation of $\gamma$ and $\omega$ restarts after each training round, ensuring that the values benefit from the improvement of the model's performance. 

Based on this mechanism, we design two strategies that modify two noise-sensitive issues summarised in the overview.
First, to solve the misguidance of the noisy anchor sample, we propose a \textbf{weighted correction strategy} to alter the labels of unreliable samples. For each sample ${\bf x} \in \mathcal{D}^N$, the weighted label ${\rm \hat y}$ is a soft label and is written as 
\begin{equation}\label{eq:weighted_label}
    {\rm {\hat y}} = (1-\omega) {\rm y^{c}} + \omega {\rm y},
\end{equation}
where ${\rm y^{c}} = {\rm Softmax} (F(x; \theta))$ and represents the prediction result of the classifier network.
Indeed, this equation only change the labels of reliable samples, \textit{i.e.}, $({\bf x},{\rm y}) \in \{({\bf x}_i, {\rm y}_i) | \omega_i \ne 1\}_{i=1}^{n}$.
Additionally, to make the alteration of labels more stable, we use the idea of moving-average. At epoch $e$, the moving-average corrected label over multiple training epochs is 
\begin{equation}\label{eq:moving_y}
    {{\rm \hat y}^e} =  \alpha {{\rm \hat y}^{(e-1)}} + (1-\alpha) {\rm \hat y}^e,
\end{equation}
where $\alpha=0.99$. Hence, the label-corrected training set can be formulated as $\{({\bf x}_i, {\rm \hat y}_i^e)\}_{i=1}^n$ at $e$-th epoch . Note that all labels are represented via a soft one-hot vector.

Second, to solve the noise tolerance properties of the momentum queue, we propose a \textbf{weighted update strategy} to solve the noise-tolerant property of the traditional momentum queue. This strategy can be simply summarized as updating this queue according to the weight in Eq. \ref{eq:eq3}. Given a sample $({\bf x}_i, {\rm \hat y}_i)$, its weight value is $\omega_i$. For the sample ${\bf x}_i$ which satisfies to $\omega_i = 1$, we update the ${\rm y}^h_i$-th queue by its feature embedding via the First-in First-out principle, where ${\rm y}^h_i$ denotes the hard label and ${\rm y}^h_i = \arg\max({\rm \hat y}_i)$. Otherwise, we update the ${\rm y}^h_i$-th queue with probability $\omega_i$. Intuitively, the weighted-update strategy avoids inserting unreliable samples into the queue, helping enhance the quality of the momentum queue.


\subsection{Stochastic feature embedding}
\label{sec:method_2}
As reported in advanced works~\cite{he2020momentum,verma2021towards}, typical CL heavily relies on sophisticated augmentation strategies and needs specify them for different datasets. We build a stochastic module to avoid manually-defined strategies.
Given a sample $x$, let ${\bf z}=f({\bf x})$ represent the output of the backbone network (\textit{i.e.}, feature extractor) and ${\bf z} \in \mathbb{R}^d$. We formulate a probability distribution $p(Q|{\bf z})$ for embedding $\bf z$ as a normal distribution, which can be written as
\begin{equation}\label{eq:sample_z}
    p(Q|{\bf z}) \sim \mathcal{N}(\mu, \sigma^2),
\end{equation}
where $\mu$ and $\sigma$ can be learned by our stochastic module, a three-layers fully-connected network. From feature embedding distribution $ p(Q|{\bf z})$, we sample an embedding ${\bf z}'$ to represent the augmented version of original feature embedding ${\bf z}$. Here, we use reparameterization trick~\cite{kingma2015variational},
\begin{equation}\label{eq:reparameter_z}
    {\bf z}' = \mu + \epsilon \cdot \sigma \quad \text{with} \quad \epsilon \sim \mathcal{N}(0, \textbf{I}).
\end{equation}
After that, the sampled feature embedding $\bf z'$ is utilized to update the momentum queue and compute contrastive learning loss. The merits of this module are 1) more complex representations are leveraged to stimulate the potential of CL, and 2) the property of stochasticity helps the model escape from memorizing the noisy signal to some degree. Module architecture has been discussed in Appx. C.1.

\subsection{Total objective}
\label{sec:method_3}
We adopt online-update strategy that alternately trains the network, alters sample labels and updates the momentum queue. At $e$-th epoch, we have a label-corrected training set $\{({\bf x}_i, {\rm \hat y}_i^e)\}_{i=1}^n$. For each sample in this set, the total training objective contains three parts.

\noindent\textbf{Classification loss}. Our proposal mitigates the effect of noisy labels by noise-tolerant representation learning, while a classification loss (\textit{e.g.} cross-entropy) is required. Due to the flexibility,
our framework can be easily integrated with prevailing LNL algorithms and leverages it for classifier learning. This loss item is written as $\mathcal{L}_{\rm LNL}$.

\noindent\textbf{Noise-tolerated contrastive loss}. For clarity, we omit the subscripts and formulate this sample as $({\bf x},{\rm \hat y})$ while $\rm q$ denotes its feature embedding and ${\rm y}^h$ represents its hard label. In our weighted momentum queue, the positive keys $\{{\rm k}^{{\rm y}^h}_1,\cdot\cdot\cdot,{\rm k}^{{\rm y}^h}_D\}$ are found according to the hard label ${\rm y}^h$. Complementarily, the remaining key points in the momentum queue are regarded as negative keys with size $[\mathrm{D} \times (C-1)]$. Note that the size of the total momentum queue is $[\mathrm{D} \times C]$. Formally, our noise-tolerated contrastive loss is summarized as
\begin{equation}\label{eq:ours}
\begin{aligned}
    \mathcal{L}_{\rm NTCL} &= - \frac{1}{\mathrm{D}} \sum\nolimits_{d = 1}^\mathrm{D} {\log \frac{{\exp (\mathrm{q} \cdot \mathrm{k}_d^{{\rm y}^h} / \tau)}}{{{L_{\rm Pos}} + {L_{\rm Neg}}}}} \quad \text{with} \\
    & L_{\rm Pos} = \sum\nolimits_{j=1}^\mathrm{D} {\exp (\mathrm{q} \cdot \mathrm{k}_j^{{\rm y}^h} / \tau)}  \quad \text{and} \quad \\
    & L_{\rm Neg} =\sum\nolimits_{c = 1}^{\{1, \cdot \cdot \cdot,C\} \backslash {{\rm y}^h}} {\sum\nolimits_{j = 1}^\mathrm{D} {\exp (\mathrm{q} \cdot \mathrm{k}_j^c / \tau)} },
\end{aligned}
\end{equation} 
where the $L_{\rm Pos}$ denotes positive keys from the same class ${\rm y}^h$ while $L_{\rm Neg}$ denotes the negative keys from other classes $\{1,\cdot \cdot \cdot,C\} \backslash {{\rm y}^h}$.

\noindent\textbf{KL regularization}. We employ the KL regularization term between the feature embedding distribution $Q$ and unit Gaussian prior $\mathcal{N}(0, \textbf{I})$ to prevent the predicted variance from collapsing to zero. The regularization can be formulated as
\begin{equation}\label{eq:kl}
    \mathcal{L}_{\rm KL} = \textbf{{\rm KL}}[p({\bf z}|Q)||\mathcal{N}(0, \textbf{I}))].
\end{equation}

The overall loss function can be formulated with two hyper-parameters $\lambda_1$ and $\lambda_2$ as
\begin{equation}\label{eq:all_l}
\mathcal{L}=\mathcal{L}_{\rm LNL} 
     + \lambda_1 \mathcal{L}_{\rm NTCL} 
     + \lambda_2 \mathcal{L}_{\rm KL}. 
\end{equation}

\begin{algorithm}[t]
	\small
	\caption{The training process of SNSCL}
	\label{alg:algorithm}
	\begin{algorithmic}[1]
	    \REQUIRE Training set $\mathcal{D}^{N}$, a reliability threshold $t \in [0,1]$, an average-moving coefficient $\alpha$, two coefficients $\lambda_1, \lambda_2$.  \\
	    \REQUIRE  Classifier network $F(\theta)$, Stochastic module $\mathcal{M}$.\\
	    \ENSURE Optimal parameters of classifier network $\theta^*$\\
	    \STATE WarmUp ($F(\theta); \mathcal{D}^N$)  \\
	    \WHILE{$e < $ MaxEpoch}
	        \STATE  Compute the loss and reliability score $\gamma$ for each sample. \\
	        \STATE  Compute the weight value $\omega$ for each sample. \hfill $\triangleright$ Eq. \ref{eq:eq3} \\
	        \STATE  Refurbish the labels with weighted-correct strategy and average-moving. \hfill $\triangleright$ {Eq. \ref{eq:weighted_label}, \ref{eq:moving_y}} \\
	        \FOR{$iter \in \{1,...,iters\}$}
	            \STATE RrandomSample a batch $\{({\bf x}_b,{\rm y}^e_b)\}_{b=1}^{\rm B}$ from the label-corrected training set, and compute loss $\mathcal{L}_{\rm LNL}$. \\ 
	            \FOR{$b \in \{1,...,{\rm B}\}$}	                
	            \STATE Sample feature embedding ${\bf z}'_b$ from the distribution $p(Q|{\bf z}_b)$.
	            \hfill  $\triangleright$ Eq.  \ref{eq:sample_z}, \ref{eq:reparameter_z}\\
	                \STATE Weighted-update the momentum queue by ${\bf z}'_b$. \\
	                \STATE Compute two losses $\mathcal{L}_{\rm KL}$, $\mathcal{L}_{\rm NTCL}$.
	                \hfill $\triangleright$ Eq. \ref{eq:ours}, \ref{eq:kl} \\ 
	            \ENDFOR
	        \STATE Update($\frac{1}{B} \sum\limits_{b=1}^{B} (\mathcal{L}_{\rm LNL} + \lambda_1 \mathcal{L}_{\rm NTCL} +  \lambda_2 \mathcal{L}_{\rm KL}); \theta^{(e)}$). \\
	        \ENDFOR
	    \ENDWHILE
	    \RETURN $\theta^*$ 
	\end{algorithmic}
\end{algorithm}

The training flowchart is shown in Fig. \ref{fig:all_fig}. Our proposed weighting strategies can be easily integrated into the typical SCL method, deriving a general LNL framework. 
The main operation is summarized in Algorithm \ref{alg:algorithm}.
Compared to typical SCL, the weighting strategies would not cause much extra computational cost.

\section{Experiments}\label{sec:experiments}

\begin{table*}[t]
\small
\centering
\caption{Comparisons with test accuracy on \textbf{symmetric} label noise. The average \textbf{best} and the \textbf{last} accuracy among three times are reported. $\uparrow$ denotes the performance improvement of \textit{SNSCL}.}
\vspace{-3mm}
\scalebox{0.85}{
\begin{tabular}{c|cccccccc}
\toprule[1.3pt]
& \multicolumn{2}{c|}{Stanford Dogs}& \multicolumn{2}{c|}{Standford Cars}& \multicolumn{2}{c|}{Aircraft}& \multicolumn{2}{c}{CUB-200-2011} \\
              & 20\%            & \multicolumn{1}{c|}{40\% } & 20\%             & \multicolumn{1}{c|}{40\% }             & 20\%          & \multicolumn{1}{c|}{40\% }          & 20\%            & 40\% \\\hline\hline
Cross-Entropy                   & 73.01 (63.82)    & 69.20 (50.45)    & 65.74 (64.08)     & 51.42 (45.62)    & 56.51 (54.67)  & 45.67 (38.89) & 64.01 (60.77)    & 54.14 (45.85)   \\ 
\multicolumn{1}{r|}{+ SNSCL}     & 76.33 (75.83)    & 75.27 (75.00)    & 83.24 (82.99)     & 76.72 (76.36)    & 76.45 (76.45)  & 70.48 (69.64) & 73.32 (72.99)    & 68.83 (68.67) \\\hline
Label Smooth~\cite{lukasik2020does}                    & 73.51 (64.42)    & 70.22 (50.97)    & 65.45 (64.24)     & 51.57 (45.19)    & 58.21 (54.73)  & 45.24 (38.01) & 64.76 (60.60)    & 54.39 (45.28)   \\
\multicolumn{1}{r|}{+ SNSCL}     & 76.85 (76.12)    & 74.64 (74.60)    & 83.21 (83.01)     & 76.07 (75.90)    & 76.24 (75.70)  & 70.36 (70.06) & 73.46 (73.09)    & 69.14 (68.64) \\\hline
Conf. Penalty~\cite{pereyra2017regularizing}                   & 73.22 (66.89)    & 68.69 (52.98)    & 64.74 (64.46)     & 48.15 (43.71)    & 56.32 (55.51)  & 43.64 (39.54) & 62.75 (61.10)    & 52.04 (45.13)   \\
\multicolumn{1}{r|}{+ SNSCL}     & 76.14 (75.73)    & 74.72 (74.49)    & 83.07 (83.00)     & 75.67 (75.38)    & 75.04 (74.23)  & 67.99 (66.85) & 73.90 (73.51)    & 68.42 (67.86)   \\\hline
GCE~\cite{zhang2018generalized}                             & 66.96 (66.93)    & 61.47 (60.32)    & 62.77 (61.23)     & 47.44 (46.13)    & 39.54 (39.24)  & 32.34 (32.28) & 58.74 (57.20)    & 49.71 (48.11)   \\
\multicolumn{1}{r|}{+ SNSCL}     & 75.99 (74.56)    & 71.68 (70.62)    & 73.78 (73.55)     & 58.11 (57.41)    & 72.67 (71.53)  & 60.19 (59.83) & 70.83 (70.56)    & 61.67 (61.46)   \\\hline
SYM~\cite{wang2019symmetric}                             & 69.20 (62.13)    & 65.76 (46.99)    & 74.65 (73.21)     & 52.83 (51.61)    & 62.29 (60.51)  & 54.36 (45.39) & 65.34 (63.60)    & 50.19 (50.15)   \\
\multicolumn{1}{r|}{+ SNSCL}     & 77.55 (77.24)    & 76.28 (76.25)    & 84.59 (83.54)     & 79.07 (78.87)    & 79.64 (79.09)  & 74.02 (73.63) & 76.67 (76.06)    & 72.71 (72.58)   \\\hline
Co-teaching~\cite{han2018co}                     & 63.71 (58.43)    & 49.15 (48.92)    & 68.60 (67.95)     & 56.92 (55.95)    & 42.55 (40.62)  & 35.21 (32.16) & 57.84 (55.98)    & 46.57 (46.22)   \\
\multicolumn{1}{r|}{+ SNSCL}     & 74.18 (73.09)    & 60.71 (58.84)    & 78.94 (78.13)     & 75.98 (75.06)    & 74.61 (74.19)  & 65.47 (63.81) & 69.77 (69.34)    & 60.59 (58.94)  
\\\hline
JoCoR~\cite{wei2020combating}                           & 66.94 (60.81)    & 49.62 (48.62)    & 69.99 (68.25)     & 57.95 (56.71)    & 61.37 (59.16)  & 52.11 (49.93) & 58.79 (57.74)    & 52.64 (49.35)   \\
\multicolumn{1}{r|}{+ SNSCL}     & 75.79 (74.99)    & 63.42 (62.84)    & 79.67 (78.77)     & 76.80 (76.21)    & 75.88 (75.16)  & 71.65 (70.67) & 71.86 (70.90)    & 64.43 (63.81)     \\\hline
MW-Net~\cite{shu2019meta}                          & 71.99 (69.20)    & 68.14 (65.17)    & 74.01 (73.88)     & 58.30 (55.81)    & 64.97 (61.84)  & 57.61 (55.90) & 67.44 (65.20)    & 58.49 (54.81)   \\
\multicolumn{1}{r|}{+ SNSCL}     & 77.49 (77.08)    & 74.92 (74.38)    & 85.96 (85.37)     & 77.76 (77.13)    & 80.08 (78.94)  & 73.55 (73.18) & 76.94 (76.24)    & 69.51 (68.83)  
\\\hline
MLC~\cite{zheng2021meta}                             & 74.08 (70.51)    & 69.44 (66.28)    & 76.02 (71.24)     & 59.44 (55.76)    & 63.81 (60.33)  & 58.11 (54.86) & 69.44 (68.19)    & 60.27 (58.49)  \\
\multicolumn{1}{r|}{+ SNSCL}     & 78.92 (78.56)    & 76.49 (78.96)    & 85.92 (84.91)     & 78.49 (77.80)    & 79.19 (78.40)  & 75.21 (74.67) & 77.58 (76.68)    & 71.54 (70.86)               \\\hline
DivideMix~\cite{li2020dividemix}                       & 79.22 (77.86)    & 77.93 (76.28)    & 78.35 (77.99)     & 62.54 (62.50)    & 80.62 (80.50)  & 66.76 (66.13) & 75.11 (74.54)    & 67.35 (66.96)    \\
\multicolumn{1}{r|}{+ SNSCL}     & 81.40 (81.16)    & 79.12 (78.91)    & 86.29 (85.94)     & 80.09 (79.51)    & 82.31 (82.03)  & 76.22 (75.67) & 78.36 (78.04)    & 73.66 (73.28)     \\\hline
\textbf{Avg. \, $\uparrow$}     & \textbf{5.88 (9.34)}      & \textbf{7.76 (15.83)}     & \textbf{12.44 (13.29)}     & \textbf{20.82 (23.06)}    & \textbf{18.60 (19.86)}  & \textbf{21.41 (24.49)} & \textbf{9.87 (11.25)}     & \textbf{12.22 (16.46)} \\
\bottomrule[1.3pt]
\end{tabular}}
\vspace{-1.5mm}
\label{tab:sym}
\end{table*}

\subsection{Implementation details}
\noindent\textbf{Noisy test benchmarks}. We introduce four typical datasets in fine-grained classification tasks and manually construct noisy labels. By a noise transition matrix $\rm T$, we change partial labels of clean datasets. Given a noise ratio $r$, for a sample $(x,y)$, the transition from clean label $y=i$ to wrong label $y=j$ can be represented by ${\rm T}_{ij} = {\rm P}(y=j|y=i)$ and ${\rm P}=r$, where $r$ is the preset noise ratio. According to the structure of $\rm T$, the noisy labels can be divided into two types: 1) \textbf{Symmetric} (random) noise. The diagonal elements of $\rm T$ are $1-r$ and the off-diagonal values are $r / (c-1)$; 2) \textbf{Asymmetric} (dependent) noise. The diagonal elements of $\rm T$ are $1-r$, and there exists another value $r$ in each row. Noise ratio $r$ is set as $r \in \{10\%,...,40\%\}$.  Illustration of the matrix $\rm T$ is shown in Appx. B.1.  

We also select two noisy dataset collected from real world (\textit{e.g.}, websites, crowdsourcing) to evaluate the effectiveness of our algorithm on real-world applications. 1) Clothing-1M~\cite{xiao2015learning} contains one million training images from 14 categories, with approximately 39.45\% noisy labels. 2) Food-101N~\cite{bossard2014food} contains 55k training images for 101 categries, with around 20\% noise ratio.

\begin{table*}[t]
\small
\centering
\caption{Comparisons with test accuracy on \textbf{asymmetric} label noise. The average \textbf{best} and the \textbf{last} accuracy among three times are reported. $\uparrow$ denotes the performance improvement of \textit{SNSCL}.}
\vspace{-3mm}
\scalebox{0.85}{
\begin{tabular}{c|cccccccc}
\toprule[1.3pt]
 &\multicolumn{2}{c|}{Stanford Dogs} &\multicolumn{2}{c|}{Standford Cars} & \multicolumn{2}{c|}{Aircraft} & \multicolumn{2}{c}{CUB-200-2011} \\
& 10\%   & \multicolumn{1}{c|}{30\% }  & 10\%   & \multicolumn{1}{c|}{30\% }   & 10\%    & \multicolumn{1}{c|}{30\% }   & 10\% & 30\%   \\\hline\hline
Cross-Entropy                            & 74.24 (71.32) & 63.76 (56.86) & 74.58 (74.57) & 58.08 (57.43) & 65.98 (62.53) & 51.10 (47.85) & 68.26 (68.00) & 56.02 (54.13) \\
\multicolumn{1}{r|}{+ SNSCL}              & 76.24 (74.88) & 64.49 (62.37) & 83.73 (83.41) & 70.04 (69.61) & 78.28 (78.22) & 65.44 (65.11) & 74.80 (74.47) & 61.48 (60.70) \\\hline
Label Smooth~\cite{lukasik2020does}                             & 74.70 (71.81) & 64.99 (57.04) & 74.28 (74.13) & 58.47 (57.80) & 65.29 (63.34) & 51.88 (47.71) & 68.78 (67.67) & 56.80 (53.69) \\
\multicolumn{1}{r|}{+ SNSCL}              & 75.84 (75.16) & 65.23 (63.69) & 84.27 (84.13) & 70.49 (70.20) & 78.67 (77.98) & 66.28 (65.56) & 75.51 (75.42) & 62.05 (61.43)  \\\hline
Conf. Penalty~\cite{pereyra2017regularizing}                            & 74.41 (72.04) & 64.50 (57.92) & 73.78 (73.67) & 56.96 (56.53) & 64.90 (63.01) & 49.38 (47.53) & 67.66 (67.62) & 54.33 (52.80) \\
\multicolumn{1}{r|}{+ SNSCL}              & 76.01 (75.62) & 67.53 (66.32) & 84.26 (83.91) & 72.23 (71.96) & 78.34 (78.01) & 66.88 (66.34) & 75.34 (74.97) & 62.69 (62.67) \\\hline
GCE~\cite{zhang2018generalized}                                      & 67.13 (66.83) & 54.53 (53.92) & 68.75 (68.71) & 60.57 (60.21) & 44.22 (44.16) & 34.18 (33.66) & 62.92 (60.77) & 50.05 (49.79) \\
\multicolumn{1}{r|}{+ SNSCL}              & 75.91 (74.63) & 68.45 (67.13) & 80.33 (80.04) & 64.64 (64.38) & 73.85 (73.89) & 64.33 (63.91) & 73.77 (73.23) & 61.37 (60.96) \\\hline
SYM~\cite{wang2019symmetric}                                      & 69.57 (66.75) & 61.61 (51.11) & 76.74 (76.18) & 58.30 (57.42) & 69.31 (67.45) & 50.23 (47.55) & 68.81 (68.00) & 52.16 (51.83) \\
\multicolumn{1}{r|}{+ SNSCL}              & 77.37 (76.64) & 74.74 (74.41) & 86.71 (86.54) & 78.98 (78.66) & 82.30 (81.46) & 69.61 (69.37) & 77.89 (77.27) & 67.43 (66.95) \\\hline
Co-teaching~\cite{han2018co}                              & 59.95 (59.77) & 50.50 (50.44) & 72.88 (72.71) & 61.02 (60.86) & 55.94 (49.85) & 45.18 (38.97) & 61.00 (60.92) & 50.06 (48.55) \\
\multicolumn{1}{r|}{+ SNSCL}              & 70.46 (70.24) & 65.83 (65.41) & 82.17 (81.63) & 66.84 (66.49) & 74.73 (74.28) & 62.17 (61.88) & 70.92 (70.63) & 64.55 (64.10) \\\hline
JoCoR~\cite{wei2020combating}                                    & 61.34 (60.11) & 53.39 (52.35) & 74.68 (73.21) & 63.54 (62.27) & 67.12 (64.99) & 52.25 (50.28) & 62.99 (61.88) & 51.70 (49.60) \\
\multicolumn{1}{r|}{+ SNSCL}              & 74.26 (72.96) & 70.40 (70.01) & 83.67 (83.28) & 71.74 (71.22) & 78.84 (78.29) & 67.50 (66.48) & 74.52 (73.97) & 66.07 (65.26)     \\\hline
MW-Net~\cite{shu2019meta}                                   & 73.68 (72.19) & 65.81 (65.19) & 76.27 (75.89) & 65.19 (63.32) & 72.76 (70.18) & 54.88 (51.80) & 67.44 (65.08) & 57.49 (56.10) \\
\multicolumn{1}{r|}{+ SNSCL}              & 78.52 (78.03) & 72.68 (72.20) & 85.73 (85.44) & 75.69 (75.28) & 80.69 (80.22) & 70.49 (69.90) & 76.07 (76.70) & 68.95 (68.26)       \\\hline
MLC~\cite{zheng2021meta}                                      & 75.84 (74.99) & 69.81 (69.03) & 77.80 (77.29) & 67.93 (67.28) & 74.40 (73.91) & 59.44 (59.00) & 68.84 (68.21) & 58.73 (58.29)\\
\multicolumn{1}{r|}{+ SNSCL}              & 79.22 (78.96) & 75.92 (75.57) & 87.05 (86.70) & 79.44 (79.21) & 82.75 (82.43) & 72.30 (71.96) & 76.91 (76.47) & 69.70 (69.24) \\\hline
DivideMix~\cite{li2020dividemix}                                & 79.39 (78.47) & 75.51 (73.67) & 79.34 (77.92) & 68.69 (68.63) & 76.57 (76.24) & 63.97 (63.28) & 72.76 (71.24) & 63.65 (62.68) \\
\multicolumn{1}{r|}{+ SNSCL}              & 81.90 (81.72) & 77.19 (77.02) & 88.18 (87.94) & 81.44 (80.96) & 84.17 (84.03) & 74.80 (74.57) & 78.92 (78.56) & 71.28 (70.83)          \\\hline
\textbf{Avg. \, $\uparrow$}  & \textbf{5.55 (6.57)} & \textbf{7.81 (10.6)} & \textbf{9.70 (9.87)} &\textbf{ 11.28 (11.62)}
                              &\textbf{13.61 (15.31)} & \textbf{16.73 (18.74)} & \textbf{8.51 (9.23) }& \textbf{10.46 (11.30)}\\
\bottomrule[1.3pt]
\end{tabular}}
\label{tab:asym}
\end{table*}

\noindent\textbf{Training settings}. The code is implemented by Pytorch 1.9.0 with single GTX 3090. For four fine-grained noisy benchmarks, the optimizer is SGD with the momentum of 0.9, while initialized learning rate is 0.001 and the weight decay is 1e-3. The number of total training epochs is both 100, and the learning rate is decayed with the factor 10 by 20 and 40 epoch.
For Clothing-1M, refers to \cite{wei2022self}, we train the classifier network for 15 epochs and use SGD with 0.9 momentum, weight decay of 5e-4. The learning rate is set as 0.002 and decayed with the factor of 10 after 10 epochs, while warm up stage is one epoch.
For Food-101N, we train the classifier network for 50 epochs and use SGD with 0.9 momentum, weight decay of 5e-4. The learning rate is set as 0.002 and decayed with the factor of 10 after 30 epochs, while warm up stage is five epoch. For all experiments, we set the training batch size as 32. In addition, we adopt a default temperature $\tau = 0.07$ for scaling. More detailed setting (\textit{e.g.}, augmentation strategies and applied backbone) can be found in Appx. B.2. 

\noindent\textbf{Hyper-parameters}. Our framework includes two hyper-parameters, \textit{i.e.}, the reliability threshold $t$ in Eq. \ref{eq:eq3} and the length of momentum queue $\mathrm{D}$ Eq. \ref{eq:ours}. For all experiments, we set $t=0.5$ and $\mathrm{D}=32$. In addition, the trade-off parameters in Eq. \ref{eq:all_l} are set as $\lambda_1=1, \lambda_2 = 0.001$. 


\begin{table*}[t]
\caption{Comparisons with test accuracy (\%) on real-world benchmarks, including Clothing-1M (the \textit{left}) and Food-101N (the \textit{right}). The \textbf{solid} results denote the improvement of our method SNSCL. The average results among five times are reported.}
\vspace{-3mm}
\centering
\makeatletter\def\@captype{table}\makeatother
\begin{minipage}{0.48\textwidth}
\small
\centering
\scalebox{0.9}{
\begin{tabular}{cc||cc}
\toprule
\multicolumn{4}{c}{Clothing-1M ($r \approx 39.5\%$)} \\ \hline\hline
Forward~\cite{patrini2017making}& 69.84         & SFT+~\cite{wei2022self}   & 75.08     \\ 
JoCoR~\cite{wei2020combating}   & 70.30         & CE                                & 64.54 \\
Joint Optim~\cite{tanaka2018joint}     & 72.23  & \cellcolor{Gray}CE + SNSCL        & \cellcolor{Gray}\textbf{73.49} \\
SL~\cite{wang2019symmetric}     & 71.02         & DivideMix~\cite{li2020dividemix}  & 74.76 \\
ELR+~\cite{liu2020early}  & 74.81               & \cellcolor{Gray}DivideMix + SNSCL & \cellcolor{Gray}\textbf{75.31} \\
\bottomrule
\end{tabular}
}
\end{minipage}
\makeatletter\def\@captype{figure}\makeatother
\begin{minipage}{0.48\textwidth}
\small
\centering
\scalebox{0.9}{
\begin{tabular}{cc||cc}
\toprule
 \multicolumn{4}{c}{Food-101N ($r \approx 20\%$)} \\\hline\hline
 CleanNet\cite{lee2017cleannet}     & 83.47     & WarPI\cite{sun2022learning}       & 85.91 \\
 MWNet\cite{shu2019meta}            & 84.72     & CE                                & 81.67 \\
 NRank\cite{sharma2020noiserank}    & 85.20     & \cellcolor{Gray}CE+SNSCL          & \cellcolor{Gray}\textbf{85.44} \\
 SMP\cite{han2019deep}              & 85.11     & DivideMix\cite{li2020dividemix}   & 85.88 \\
 PLC\cite{zhang2021learning}        & 85.28     & \cellcolor{Gray}DivideMix+SNSCL   & \cellcolor{Gray}\textbf{86.40} \\
\bottomrule
\end{tabular}}
\end{minipage}
\label{tab:cloth_cifar}
\vspace{-2mm}
\end{table*}

\begin{figure*}[t]
\centering
\includegraphics[width=0.93\linewidth]{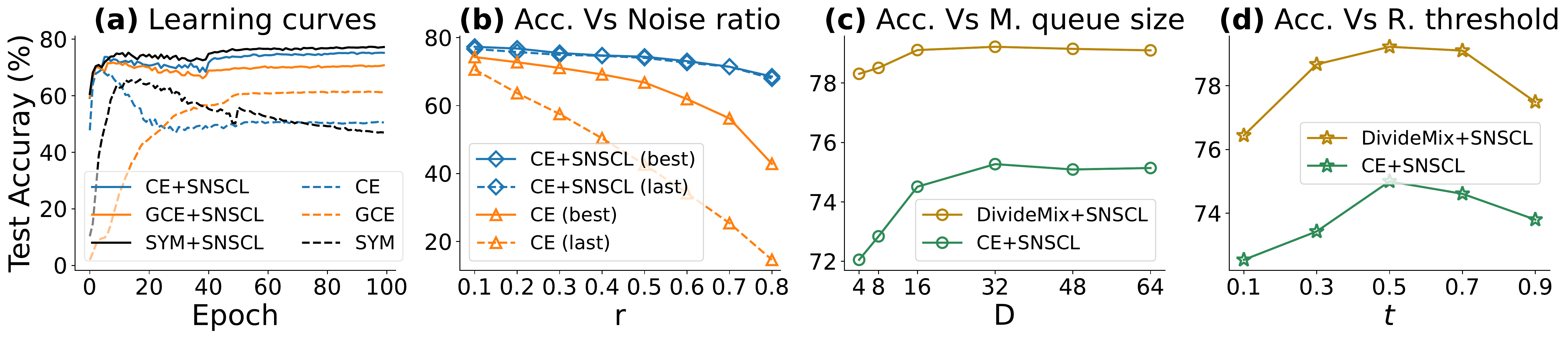}
\vspace{-3mm}
\caption{More analyses on Stanford dogs with 40\% symmetric label noise from four perspectives .}
\vspace{-3mm}
\label{fig:analysis}
\end{figure*}

\subsection{Comparison with state-of-the-arts}
\noindent\textbf{Baselines}. We evaluate the effectiveness of our method by adding the proposal into current LNL algorithm and compare the improvements on LNL-FG task. The basic methods we compared include CE, Label Smooth~\cite{lukasik2020does}, Confidence Penalty~\cite{pereyra2017regularizing}, Co-teaching~\cite{han2018co}, JoCoR~\cite{wei2020combating}, DivideMix~\cite{li2020dividemix}, SYM~\cite{wang2019symmetric}, GCE~\cite{zhang2018generalized}, MW-Net~\cite{shu2019meta}, and MLC~\cite{zheng2021meta}. Settings about these methods are shown in Appx. B.3.

\noindent\textbf{Results on four fine-grained benchmarks}. We compare 10 algorithms and attain significant improvement of top-1 testing accuracy (\%) on four fine-grained benchmarks. We show the results in Tab. \ref{tab:sym} and Tab. \ref{tab:asym}, where we test symmetric and asymmetric noise types.
To demonstrate the effectiveness of our method, we give experimental comparisons from two aspects.
1) \textbf{Improvements on top-1 test accuracy}. Overall, our method SNSCL achieves consistent improvement in all noisy conditions. The average minimal improvement is 5.55\% in Stanford Dogs with 10\% asymmetric noise, and the maximum is 21.41\% in Aircraft with 40\% symmetric noise. 2) \textbf{Mitigating overfitting on fine-grained sets}. In these tables, we report the best accuracy and the last epoch's accuracy. It is noteworthy that the investigated methods mainly overfit on these benchmarks (\textit{i.e.}, the accuracy attaches a peak and then drops gradually, causing a great gap between these two values). However, SNSCL mitigates overfitting and maintains more stable learning curves.

\noindent\textbf{Results on real-world noisy datasets}. To evaluate the effectiveness of SNSCL on real-world applications, we conduct experiments on two datasets which are collected from websites. 1) \textbf{Clothing-1M}, the comparison results is shown in Tab. \ref{tab:cloth_cifar}. We select cross-entropy, GCE and DivideMix as the basic methods and integrate SNSCL with them. Obviously, the combination \textit{DivideMix+SNSCL} outperforms the state-of-the-art method SFT+ by 0.23\% top-1 test accuracy. Moreover, in contrast with the bases, SNSCL achieves remarkable improvements by 8.95\% and 0.55\%, respectively.
2) \textbf{Food-101N}, the comparison results is shown in Tab. \ref{tab:cloth_cifar}. Compared to basic methods, SNSCL brings significantly improvement by 3.77\% and 0.52\%, respectively. These results demonstrate the effectiveness of our methods in real-world applications.

\noindent\textbf{Results on CIFAR-10 \& 100}. SNSCL plays against with negative effects of label noise by enhancing distinguishable representation, which is also suitable for generic noisy classification to some degree. We conduct experiments on synthetic noisy CIFAR-10 \& 100 and show the results in Appx. C.2. Overall, the performance of tested methods achieve non-trivial improvements by combining with SNSCL.

\subsection{More analysis}

\begin{table*}[t]
\small
\centering
\caption{Compared with previous contrastive-based methods on four noisy benchmarks.}
\vspace{-4mm}
\scalebox{0.93}{
\begin{tabular}{c|ccc|ccc|cc|cc}
\toprule
Dataset     & \multicolumn{3}{c|}{CIFAR-10}      & \multicolumn{3}{c|}{CIFAR-100}      & \multicolumn{2}{c|}{Stanford Dogs}   & \multicolumn{2}{c}{CUB-200-2011}   \\
Noise Type       & S. 50\% & S. 80\%  & A. 40\%       & S. 50\%  & S. 80\%       & A. 40\%    & S. 40\%      & A. 30\%    & S. 40\%      & A. 30\%           \\ \hline\hline
MoPro ~\cite{li2021mopro}          & \textbf{95.6}  & 90.1          & 93.0 & \textbf{74.9}   & 61.9    & 73.0   & 78.41\scriptsize{$\pm\text{0.1}$} & 74.39\scriptsize{$\pm\text{0.2}$}   & 73.23\scriptsize{$\pm\text{0.2}$} & 68.58\scriptsize{$\pm\text{0.4}$} \\
Sel-CL+~\cite{li2022selective}     & 93.9           & 89.2          & 93.4          & 72.4   & 59.6    & 74.2   & 77.92\scriptsize{$\pm\text{0.3}$} & 75.29\scriptsize{$\pm\text{0.2}$}   & 73.01\scriptsize{$\pm\text{0.1}$} & 70.47\scriptsize{$\pm\text{0.1}$}               \\
\textbf{Ours}                      & 95.2           & \textbf{91.7} & \textbf{94.9} & 74.7   & \textbf{64.3}  & \textbf{75.1}  & \textbf{79.13\scriptsize{$\pm\text{0.2}$}} & \textbf{77.20\scriptsize{$\pm\text{0.1}$}} & \textbf{73.67\scriptsize{$\pm\text{0.3}$}} & \textbf{71.28\scriptsize{$\pm\text{0.2}$}}\\
\bottomrule
\end{tabular}
}
\vspace{-2mm}
\label{tab:sel_cl}
\end{table*}

\noindent\textbf{Effectiveness}. Our algorithm exhibits the superior effectiveness in two aspects. 1) we plot the curve of test accuracy in Fig. \ref{fig:analysis}(a). It is clear that the accuracy of CE rises dramatically to a peak and gradually decreases, indicating overfitting to noise under fine-grained datasets. For SNSCL, the testing curve is relatively stable and results in good generalization performance. 2) we test the noise ratios with a wide range of $r \in \{10\%,\cdot\cdot\cdot,80\%\}$ and record the best and the last top-1 testing accuracy. As shown in the scatter Fig. \ref{fig:analysis}(b), SNSCL can mitigate reasonable discriminability for a high noise ratio (68\% top-1 accuracy for symmetric 80\% label noise). 
Meanwhile, more train curves with varying noise ratios are shown in Appx. C.4.

\noindent\textbf{Sensibility}.  We explore the effect of two essential hyper-parameters in our method. 1) \textbf{The momentum queue size $\rm{D}$}. The batch size or momentum queue size is the key point in contrastive learning, and thus we set $\rm{D} \in \{4,8,16,32,48,64\}$ to explore its influence on our framework. The results are shown in Fig. \ref{fig:analysis}(c). As the size $\rm{D}$ reaches a certain amount, the performance will not increase. Thus, we set a suitable yet effective value $\rm{D}=32$. 2) \textbf{The reliability threshold $t$}. This threshold in the weight-aware mechanism deeply affects the subsequent two weighted strategies. We adjust its value from the space $\{0.1,0.3,0.5,0.7,0.9\}$ and plot the results in Fig. \ref{fig:analysis}(d). The best performance is attained on two conditions when $t=0.5$. Therefore, we set the reliability threshold as $0.5$.


\noindent\textbf{Compared with contrastive-based LNL methods}. We conduct experiments to compare our method (DivideMix + SNSCL) with MoPro \cite{li2021mopro} and Sel-CL+ \cite{li2022selective}, two LNL methods based on contrastive learning. Detailed discussions about these methods can be found in Related works. 

Tab. \ref{tab:sel_cl} reports the comparison results with top-1 test accuracy on four benchmarks. Our method outperforms Sel-CL+ and MoPro in most noisy settings. As the noise ratio arises, the achievements of SNSCL are more remarkable. Compared to Sel-CL+ while the performance is improved by 2.5\% on CIFAR-10 80\% symmetric noise, and 4.7\% on CIFAR-100 80\% symmetric noise. Under four LNL-FG settings, our methods consistently outperform other methods. Compared to Sel-CL+, the improvement is roughly 2\% on Stanford Dogs 30\% asymmetric noise. 

\noindent\textbf{Discussion about the stochastic module}. We conduct comparison experiments with traditional augmentation strategies to verify the ability of representation enhancement of stochastic module. As shown in Tab. \ref{tab:aug_vs_}, our proposed module exhibits greater performance under noisy conditions. Compared to strong augmentation, the average improvement is more than 1\%. Besides, the combination of our stochastic module and strong augmentation does not bring improvements. Thus, we do not adopt strong augmentation strategies in our training framework.

\noindent\textbf{Ablation study}. In our proposed SNSCL, there mainly exists three components, weighted-correction and weighted-update strategy in a weighted-aware mechanism and a stochastic module. We conduct the ablation study on two benchmarks to evaluate the effectiveness of each component and show the results in Tab. \ref{tab:ablation}. Under the settings of Stanford dogs with 40\% symmetric noisy labels, the combination of three components improves the performance of CE by more than 6\% and the effect of DivideMix by 3\% respectively, while all components bring some positive effects. Meanwhile, due to the noise-sensitivity of SCL, integrating SCL into CE brings performance degradation instead.   
To some extent, these results demonstrate the effectiveness of each part of our method. 

\noindent\textbf{Visualization}. To demonstrate the distinguishable classes are learnt by our proposed SNSCL, we leverage t-SNE\cite{van2008visualizing} to visualize the feature embeddings on the testing sets of CIFAR-10 \& CIFAR-100. The results are shown in Appx. C.3, verifying the improvement of SNSCL on representation learning under noisy conditions.

\begin{table}[t]
\centering
\small
\caption{Compared stochastic module (Ours) to weak (W.) and strong (S.) augmentation under 40\% symmetric label noise.}
\vspace{-3mm}
\scalebox{0.93}{
\begin{tabular}{c|ccc}
\toprule
Strategies     & Stanford Dogs & CUB-200-2011 & Aircraft \\\hline\hline
W. aug.      & 73.24\scriptsize{$\pm$0.2}       & 66.48\scriptsize{$\pm$0.2}   & 68.74\scriptsize{$\pm$0.5} \\
S. aug.    & 74.02\scriptsize{$\pm$0.3}         & 67.26\scriptsize{{$\pm$0.4}}  & 70.19\scriptsize{$\pm$0.2}\\ 
\cellcolor{Gray}Ours   & \cellcolor{Gray}\textbf{75.27\scriptsize{$\pm$0.2}}    & \cellcolor{Gray}69.09{\scriptsize{$\pm$0.4}}  & \cellcolor{Gray}\textbf{70.48\scriptsize{$\pm$0.3}}\\
\cellcolor{Gray}Ours + S. aug.  & \cellcolor{Gray}75.13\scriptsize{$\pm$0.2}    & \cellcolor{Gray}\textbf{69.31\scriptsize{$\pm$0.2}} & \cellcolor{Gray}70.19\scriptsize{$\pm$0.3}\\
\bottomrule
\end{tabular}
}
\vspace{-1mm}
\label{tab:aug_vs_}
\end{table}

\begin{table}[t]
\centering
\small
\caption{\textbf{Ablation study} about the effectiveness of each component under 40\% \textit{symm.} label noise.}
\vspace{-3mm}
\scalebox{0.94}{
\begin{tabular}{l|cc}
\toprule
                        & Stanford Dogs & CUB-200-2011 \\\hline\hline
\multicolumn{1}{c|}{CE} & 69.20 (50.45)                & 54.14 (45.85)   \\\hline
 CE + SCL               & 68.49 (54.77)                & 53.30 (45.92) \\             
\cellcolor{Gray}CE + SNSCL  & \cellcolor{Gray}\textbf{75.27 (75.00)}                &\cellcolor{Gray} \textbf{68.83 (68.67) }   \\
\quad w/o Weight corr.   & 70.91\scriptsize{$\pm$0.6}   & 62.71\scriptsize{$\pm$0.5}   \\
\quad w/o Weight update & 73.45\scriptsize{$\pm$0.3}   & 65.29\scriptsize{$\pm$0.4}             \\
\quad w/o Stoc. module  & 74.11\scriptsize{$\pm$0.3}   & 67.44\scriptsize{$\pm$0.3} \\\hline\hline
\multicolumn{1}{c|}{DivideMix} & 77.93 (76.28)              & 67.35 (66.96)    \\\hline
DivideMix + SCL               & 78.20 (77.89)                 & 70.28 (70.02) \\
\cellcolor{Gray}DivideMix + SNSCL  & \cellcolor{Gray} \textbf{79.12 (78.91)}             & \cellcolor{Gray} \textbf{73.66 (73.28)}\\
\quad w/o Weight corr.          & 78.30\scriptsize{$\pm$0.2} & 70.41\scriptsize{$\pm$0.3}  \\
\quad w/o Weight update        & 78.52\scriptsize{$\pm$0.1} & 72.59\scriptsize{$\pm$0.2}  \\
\quad w/o Stoc. module         & 78.85\scriptsize{$\pm$0.1} & 73.06\scriptsize{$\pm$0.1}  \\
\bottomrule
\end{tabular}
\label{tab:ablation}
}
\vspace{-3mm}
\end{table}

\section{Conclusion}\label{sec:conclusions}
In this work, we propose a novel task called LNL-FG, posing a more challenging noisy scenario to learning with noisy labels. For this, we design a general framework called SNSCL. SNSCL contains a noise-tolerated contrastive loss and a stochastic module. Compared with typical SCL, our contrastive learning framework incorporates a weight-aware mechanism which corrects noisy labels and selectively update momentum queue lists. Besides, we propose a stochastic module for feature transformation, generating the probabilistic distribution of feature embeddings. We achieve greater representation ability by sampling transformed embedding from this distribution. SNSCL is applicable to prevailing LNL methods and further improves their generalization performance on LNL-FG. Extensive experiments and analysis verify the effectiveness of our method.

\vspace{-2.5mm}
\section*{Acknowledgements}
\vspace{-1.5mm}
Haoliang Sun was supported by Natural Science Foundation of China (No. 62106129), China Postdoctoral Science Foundation (No. 2021TQ0195, 2021M701984) and Natural Science Foundation of Shandong Province (No. ZR2021QF053).
Yilong Yin was supported by Major Basic Research Project of Natural
Science Foundation of Shandong Province (No. ZR2021ZD15) and Natural Science Foundation of China (No. 62176139).
Lei Feng was supported by the National Natural Science Foundation of China (No. 62106028), Chongqing Overseas Chinese Entrepreneurship and Innovation Support Program, and CAAI-Huawei MindSpore Open Fund.

{\small
\bibliographystyle{ieee_fullname}
\bibliography{main}

\begin{thebibliography}{10}\itemsep=-1pt

\bibitem{bai2021me}
Yingbin Bai and Tongliang Liu.
\newblock Me-momentum: Extracting hard confident examples from noisily labeled
  data.
\newblock In {\em ICCV}, 2021.

\bibitem{bossard2014food}
Lukas Bossard, Matthieu Guillaumin, and Luc Van~Gool.
\newblock Food-101--mining discriminative components with random forests.
\newblock In {\em ECCV}, 2014.

\bibitem{bukchin2021fine}
Guy Bukchin, Eli Schwartz, Kate Saenko, Ori Shahar, Rogerio Feris, Raja Giryes,
  and Leonid Karlinsky.
\newblock Fine-grained angular contrastive learning with coarse labels.
\newblock In {\em CVPR}, 2021.

\bibitem{chen2020simple}
Ting Chen, Simon Kornblith, Mohammad Norouzi, and Geoffrey Hinton.
\newblock A simple framework for contrastive learning of visual
  representations.
\newblock In {\em ICML}, 2020.

\bibitem{chen2020big}
Ting Chen, Simon Kornblith, Kevin Swersky, Mohammad Norouzi, and Geoffrey~E
  Hinton.
\newblock Big self-supervised models are strong semi-supervised learners.
\newblock In {\em NeurIPS}, 2020.

\bibitem{grill2020bootstrap}
Jean-Bastien Grill, Florian Strub, Florent Altch{\'e}, Corentin Tallec, Pierre
  Richemond, Elena Buchatskaya, Carl Doersch, Bernardo Avila~Pires, Zhaohan
  Guo, Mohammad Gheshlaghi~Azar, et~al.
\newblock Bootstrap your own latent-a new approach to self-supervised learning.
\newblock In {\em NeurIPS}, 2020.

\bibitem{han2018co}
Bo Han, Quanming Yao, Xingrui Yu, Gang Niu, Miao Xu, Weihua Hu, Ivor Tsang, and
  Masashi Sugiyama.
\newblock Co-teaching: Robust training of deep neural networks with extremely
  noisy labels.
\newblock In {\em NeurIPS}, 2018.

\bibitem{han2019deep}
Jiangfan Han, Ping Luo, and Xiaogang Wang.
\newblock Deep self-learning from noisy labels.
\newblock In {\em ICCV}, 2019.

\bibitem{he2020momentum}
Kaiming He, Haoqi Fan, Yuxin Wu, Saining Xie, and Ross Girshick.
\newblock Momentum contrast for unsupervised visual representation learning.
\newblock In {\em CVPR}, 2020.

\bibitem{jaiswal2020survey}
Ashish Jaiswal, Ashwin~Ramesh Babu, Mohammad~Zaki Zadeh, Debapriya Banerjee,
  and Fillia Makedon.
\newblock A survey on contrastive self-supervised learning.
\newblock {\em Technologies}, 9, 2020.

\bibitem{khosla2020supervised}
Prannay Khosla, Piotr Teterwak, Chen Wang, Aaron Sarna, Yonglong Tian, Phillip
  Isola, Aaron Maschinot, Ce Liu, and Dilip Krishnan.
\newblock Supervised contrastive learning.
\newblock In {\em NeurIPS}, 2020.

\bibitem{kingma2015variational}
Durk~P Kingma, Tim Salimans, and Max Welling.
\newblock Variational dropout and the local reparameterization trick.
\newblock In {\em NeurIPS}, 2015.

\bibitem{lee2017cleannet}
Kuang-Huei Lee, Xiaodong He, Lei Zhang, and Linjun Yang.
\newblock Cleannet: Transfer learning for scalable image classifier training
  with label noise.
\newblock In {\em CVPR}, 2018.

\bibitem{li2020dividemix}
Junnan Li, Richard Socher, and Steven~CH Hoi.
\newblock Dividemix: Learning with noisy labels as semi-supervised learning.
\newblock In {\em ICLR}, 2020.

\bibitem{li2021mopro}
Junnan Li, Caiming Xiong, and Steven~CH Hoi.
\newblock Mopro: Webly supervised learning with momentum prototypes.
\newblock In {\em ICLR}, 2021.

\bibitem{li2022exploring}
Mingsong Li, Yikun Liu, Guangkuo Xue, Yuwen Huang, and Gongping Yang.
\newblock Exploring the relationship between center and neighborhoods: Central
  vector oriented self-similarity network for hyperspectral image
  classification.
\newblock {\em IEEE Trans. Circuits Syst. Video Technol.}, 2022.

\bibitem{li2022selective}
Shikun Li, Xiaobo Xia, Shiming Ge, and Tongliang Liu.
\newblock Selective-supervised contrastive learning with noisy labels.
\newblock In {\em CVPR}, 2022.

\bibitem{liu2020early}
Sheng Liu, Jonathan Niles-Weed, Narges Razavian, and Carlos Fernandez-Granda.
\newblock Early-learning regularization prevents memorization of noisy labels.
\newblock In {\em NeurIPS}, 2020.

\bibitem{liu2020peer}
Yang Liu and Hongyi Guo.
\newblock Peer loss functions: Learning from noisy labels without knowing noise
  rates.
\newblock In {\em ICML}, 2020.

\bibitem{long2008random}
Philip~M Long and Rocco~A Servedio.
\newblock Random classification noise defeats all convex potential boosters.
\newblock In {\em ICML}, 2008.

\bibitem{lukasik2020does}
Michal Lukasik, Srinadh Bhojanapalli, Aditya Menon, and Sanjiv Kumar.
\newblock Does label smoothing mitigate label noise?
\newblock In {\em ICML}, 2020.

\bibitem{ma2020normalized}
Xingjun Ma, Hanxun Huang, Yisen Wang, Simone Romano, Sarah Erfani, and James
  Bailey.
\newblock Normalized loss functions for deep learning with noisy labels.
\newblock In {\em ICML}, 2020.

\bibitem{natarajan2013learning}
Nagarajan Natarajan, Inderjit~S Dhillon, Pradeep~K Ravikumar, and Ambuj Tewari.
\newblock Learning with noisy labels.
\newblock In {\em NeurIPS}, 2013.

\bibitem{nguyen2019self}
Duc~Tam Nguyen, Chaithanya~Kumar Mummadi, Thi Phuong~Nhung Ngo, Thi Hoai~Phuong
  Nguyen, Laura Beggel, and Thomas Brox.
\newblock Self: Learning to filter noisy labels with self-ensembling.
\newblock In {\em ICLR}, 2020.

\bibitem{oord2018representation}
Aaron van~den Oord, Yazhe Li, and Oriol Vinyals.
\newblock Representation learning with contrastive predictive coding.
\newblock {\em arXiv preprint arXiv:1807.03748}, 2018.

\bibitem{park2020contrastive}
Taesung Park, Alexei~A Efros, Richard Zhang, and Jun-Yan Zhu.
\newblock Contrastive learning for unpaired image-to-image translation.
\newblock In {\em ECCV}, 2020.

\bibitem{patrini2017making}
Giorgio Patrini, Alessandro Rozza, Aditya Krishna~Menon, Richard Nock, and
  Lizhen Qu.
\newblock Making deep neural networks robust to label noise: A loss correction
  approach.
\newblock In {\em CVPR}, 2017.

\bibitem{pereyra2017regularizing}
Gabriel Pereyra, George Tucker, Jan Chorowski, {\L}ukasz Kaiser, and Geoffrey
  Hinton.
\newblock Regularizing neural networks by penalizing confident output
  distributions.
\newblock In {\em ICLR}, 2017.

\bibitem{ren2018learning}
Mengye Ren, Wenyuan Zeng, Bin Yang, and Raquel Urtasun.
\newblock Learning to reweight examples for robust deep learning.
\newblock In {\em ICML}, 2018.

\bibitem{sharma2020noiserank}
Karishma Sharma, Pinar Donmez, Enming Luo, Yan Liu, and I~Zeki Yalniz.
\newblock Noiserank: Unsupervised label noise reduction with dependence models.
\newblock In {\em ECCV}, 2020.

\bibitem{shu2019meta}
Jun Shu, Qi Xie, Lixuan Yi, Qian Zhao, Sanping Zhou, Zongben Xu, and Deyu Meng.
\newblock Meta-weight-net: Learning an explicit mapping for sample weighting.
\newblock In {\em NeurIPS}, 2019.

\bibitem{song2019selfie}
Hwanjun Song, Minseok Kim, and Jae-Gil Lee.
\newblock Selfie: Refurbishing unclean samples for robust deep learning.
\newblock In {\em ICML}, 2019.

\bibitem{sun2022learning}
Haoliang Sun, Chenhui Guo, Qi Wei, Zhongyi Han, and Yilong Yin.
\newblock Learning to rectify for robust learning with noisy labels.
\newblock {\em Pattern Recognition}, 2022.

\bibitem{sun2017direct}
Haoliang Sun, Xiantong Zhen, Chris Bailey, Parham Rasoulinejad, Yilong Yin, and
  Shuo Li.
\newblock Direct estimation of spinal cobb angles by structured multi-output
  regression.
\newblock In {\em ICIP}, 2017.

\bibitem{tanaka2018joint}
Daiki Tanaka, Daiki Ikami, Toshihiko Yamasaki, and Kiyoharu Aizawa.
\newblock Joint optimization framework for learning with noisy labels.
\newblock In {\em CVPR}, 2018.

\bibitem{tewari2007consistency}
Ambuj Tewari and Peter~L Bartlett.
\newblock On the consistency of multiclass classification methods.
\newblock {\em Journal of Machine Learning Research}, 2007.

\bibitem{van2008visualizing}
Laurens Van~der Maaten and Geoffrey Hinton.
\newblock Visualizing data using t-sne.
\newblock {\em Journal of machine learning research}, 2008.

\bibitem{verma2021towards}
Vikas Verma, Thang Luong, Kenji Kawaguchi, Hieu Pham, and Quoc Le.
\newblock Towards domain-agnostic contrastive learning.
\newblock In {\em ICML}, 2021.

\bibitem{wang2021self}
Ximei Wang, Jinghan Gao, Mingsheng Long, and Jianmin Wang.
\newblock Self-tuning for data-efficient deep learning.
\newblock In {\em ICML}, 2021.

\bibitem{wang2021proselflc}
Xinshao Wang, Yang Hua, Elyor Kodirov, David~A Clifton, and Neil~M Robertson.
\newblock Proselflc: Progressive self label correction for training robust deep
  neural networks.
\newblock In {\em CVPR}, 2021.

\bibitem{wang2019symmetric}
Yisen Wang, Xingjun Ma, Zaiyi Chen, Yuan Luo, Jinfeng Yi, and James Bailey.
\newblock Symmetric cross entropy for robust learning with noisy labels.
\newblock In {\em ICCV}, 2019.

\bibitem{wei2020combating}
Hongxin Wei, Lei Feng, Xiangyu Chen, and Bo An.
\newblock Combating noisy labels by agreement: A joint training method with
  co-regularization.
\newblock In {\em CVPR}, 2020.

\bibitem{wei2022self}
Qi Wei, Haoliang Sun, Xiankai Lu, and Yilong Yin.
\newblock Self-filtering: A noise-aware sample selection for label noise with
  confidence penalization.
\newblock In {\em ECCV}, 2022.

\bibitem{wu2020learning}
Yichen Wu, Jun Shu, Qi Xie, Qian Zhao, and Deyu Meng.
\newblock Learning to purify noisy labels via meta soft label corrector.
\newblock In {\em AAAI}, 2021.

\bibitem{xiao2015learning}
Tong Xiao, Tian Xia, Yi Yang, Chang Huang, and Xiaogang Wang.
\newblock Learning from massive noisy labeled data for image classification.
\newblock In {\em CVPR}, 2015.

\bibitem{xu2019l_dmi}
Yilun Xu, Peng Cao, Yuqing Kong, and Yizhou Wang.
\newblock L\_dmi: An information-theoretic noise-robust loss function.
\newblock In {\em NeurIPS}, 2019.

\bibitem{zhang2021learning}
Yivan Zhang, Gang Niu, and Masashi Sugiyama.
\newblock Learning noise transition matrix from only noisy labels via total
  variation regularization.
\newblock In {\em ICML}, 2021.

\bibitem{zhang2018generalized}
Zhilu Zhang and Mert~R Sabuncu.
\newblock Generalized cross entropy loss for training deep neural networks with
  noisy labels.
\newblock In {\em NeurIPS}, 2018.

\bibitem{zheng2021meta}
Guoqing Zheng, Ahmed~Hassan Awadallah, and Susan Dumais.
\newblock Meta label correction for noisy label learning.
\newblock In {\em AAAI}, 2021.

\end{thebibliography}
}

\end{document}